\documentclass[journal]{IEEEtran}

\usepackage{amsmath,amsfonts,amssymb}
\usepackage{algorithm}
\usepackage{algpseudocode}
\usepackage{array}
\usepackage{booktabs}
\usepackage{cite}
\usepackage{graphicx}
\usepackage{makecell}
\usepackage{multirow}
\usepackage{stfloats}
\usepackage{textcomp}
\usepackage{url}

\begin{document}
\raggedbottom{}

\title{DriftingVLA: Native One-Step Vision-Language-Action Generation via Per-Dimension Temporal Drifting}
\author{
Yuxuan Gao$^{1,*}$,
Shiqi Zhang$^{1,*}$,
Yedong Shen$^{1}$,
Yifan Duan$^{2}$,
Wenhao Yu$^{1}$,
Xin Zhang$^{1}$,
Siyuan Cao$^{1}$,
Jiajun Deng$^{1}$,
Yanyong Zhang$^{1,\dagger}$~\IEEEmembership{Fellow,~IEEE}
\thanks{$^{*}$These authors contributed equally.}
\thanks{$^{\dagger}$Corresponding author.}
\thanks{$^{1}$University of Science and Technology of China,
Hefei 230026, China. Emails: \{yuxuangao, zhangshiqi\_1127,
sydong2002, wenhaoyu, xzhangl2, caosiyuan\}@mail.ustc.edu.cn,
\{dengjj, yanyongz\}@ustc.edu.cn}
\thanks{$^{2}$LYNSENSE, Shanghai, China. Email: duanyifan@lynsense.net}
}
\maketitle

\begin{figure*}[t]
    \centering
    \includegraphics[width=0.98\textwidth]{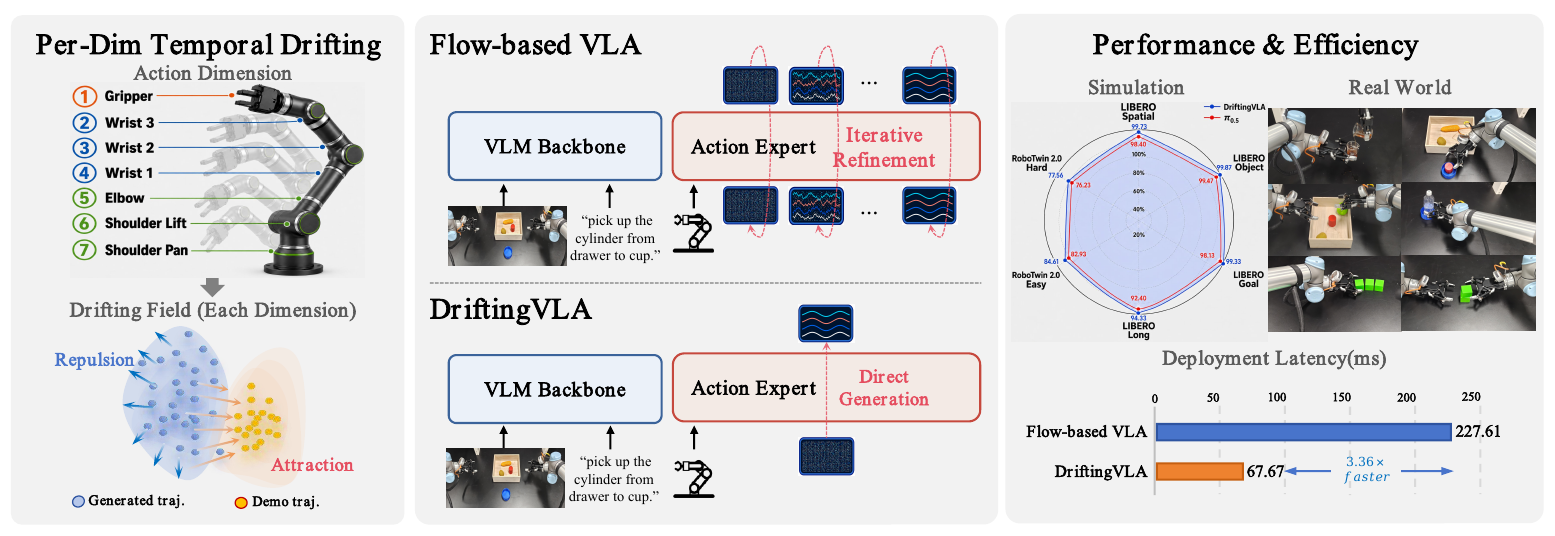}
    \caption{Overview of DriftingVLA. Conventional flow-based VLAs generate an action chunk through iterative action-expert evaluations along a numerical integration trajectory. DriftingVLA instead performs distribution refinement during training via Per-Dimension Temporal Drifting (PDTD) and deploys a direct conditional generator that requires only one action-expert evaluation per action chunk. The right panel summarizes simulation and real-world performance together with deployment-time efficiency.}
    \label{fig:overview}
\end{figure*}



\begin{abstract}

Conventional flow-based vision-language-action (VLA) models support expressive continuous action generation but rely on multi-step refinement to produce each action chunk, increasing latency in online robot control. To address this issue, we introduce DriftingVLA, a native one-step VLA that generates a complete action chunk with a single action-expert forward pass. Rather than learning a flow field that requires iterative integration at inference, DriftingVLA uses a distribution-drifting objective to learn a direct noise-to-action-chunk mapping for one-step deployment. Since robot action dimensions carry distinct control semantics and distributional characteristics, we further introduce Per-Dimension Temporal Drifting (PDTD). PDTD treats the complete temporal trajectory of each action dimension as a separate drifting unit, enabling finer-grained modeling and shaping of dimension-specific action distributions. This per-dimension decomposition applies only to the training objective; the shared VLA model still generates the complete action chunk jointly, thereby preserving cross-dimensional dependencies. DriftingVLA achieves 98.32\% success on LIBERO, 81.09\% on RoboTwin 2.0, and 77.67\% across six real-world single- and dual-arm tasks, outperforming the evaluated multi-step flow policy and one-step VLA baselines. Native one-step deployment also accelerates action-chunk generation by 3.36$\times$, eliminating iterative refinement without sacrificing control performance.

\end{abstract}


\section{Introduction}\label{sec:introduction}

Vision-language-action (VLA) models have emerged as a promising paradigm for general-purpose robotic manipulation by combining pretrained multimodal representations with end-to-end continuous control~\cite{palme2023,rt1_2023,rt2_2023,openx2024,octo2024,openvla2025}. Recent models further improve action expressiveness by pairing a large vision-language backbone with a generative continuous-action head~\cite{openvlaoft2025,rdt1b2024,cogact2024,smolvla2025,grootn1_2025}. In particular, $\pi_0$~\cite{pi0_2024} and $\pi_{0.5}$~\cite{pi05_2024} use flow matching to generate action chunks, enabling the policy to model complex and multimodal robot behaviors while retaining the semantic and perceptual priors of a pretrained VLM backbone. This design, however, introduces a deployment-time cost: each action chunk is recovered by numerically integrating a learned vector field, requiring multiple sequential evaluations of the action expert. Although the multimodal prefix can be cached across the integration trajectory, the repeated action-expert computation remains directly on the robot-time critical path.

This latency has motivated a growing body of work on faster generative action policies~\cite{prasad2024consistency,wang2024onestepdiffusionpolicyfast,falcon2025,smolvla2025,onestepflowpolicy2026}. Some approaches reduce execution latency while retaining an iterative generator, for example by overlapping action generation with execution or prioritizing near-term actions during flow sampling~\cite{rtc_2025,faster_2026,asyncvla2025}. More directly, recent one-step VLA methods modify the training or action-generation formulation~\cite{letitbesimple2026,invertibleadapter2026,mmact2025,onestepflowpolicy2026}. SnapFlow~\cite{snapflow2026} progressively distills a multi-step flow trajectory into a shortcut predictor, while MeanFlowVLA~\cite{meanflowvla_2026} adapts the MeanFlow formulation~\cite{meanflow2024} to predict an interval-averaged velocity suitable for one-step action generation. These methods demonstrate that flow-based VLAs can be accelerated substantially, but their one-step generators remain derived from the transport process inherited from flow matching. This motivates a different formulation: adapting a pretrained flow-based VLA into a native one-step conditional generator without relying on an inference-time transport trajectory.

Distribution drifting provides a natural foundation for this formulation. Drifting Models~\cite{deng2026generative} move distribution refinement from inference-time sample evolution to optimization-time generator learning. Rather than learning a local denoising or transport field that must be repeatedly evaluated after deployment, a drifting field specifies distribution-level corrections during training, and these corrections are progressively absorbed into the generator parameters. Drift-Based Policy Optimization (DBPO)~\cite{gao2026drift} introduced a native one-step policy learning framework for robotic control. Its generative component, Drift-Based Policy (DBP), applies distribution drifting in action space using multiple sibling action hypotheses during training while retaining single-sample, one-evaluation deployment. Extending DBP to pretrained flow-based VLAs, however, is not a direct substitution and raises two coupled design challenges. First, the generative interface must be adapted: the action expert of a pretrained flow VLA is specialized for local velocity prediction, whereas native drifting requires a direct noise-to-action generator. Second, the distributional geometry must be reconsidered: a VLA predicts a structured, temporally extended action chunk, so the organization of its temporal and action dimensions directly determines the geometry used to construct drifting updates.

Specifically, an action chunk is naturally represented as a tensor $\mathbf A\in\mathbb R^{H\times D}$, where the temporal horizon $H$ and action dimension $D$ form two semantically different axes~\cite{act2023,diffusionpolicy2023,openvlaoft2025,baku2024,fast_2025,dp3_2024}. The Chunk-wise and Step-wise drifting organizations considered in DBP expose complementary limitations when applied to this structure. Chunk-wise drifting flattens the entire $H\times D$ tensor into one group. It preserves the complete temporal horizon, but translation, rotation, gripper, and other heterogeneous action channels must share the same distance scale, neighborhood statistics, and drifting geometry. Step-wise drifting instead treats each temporal slice independently. This reduces the scope of each group, but fragments the complete temporal evolution of every action channel across independently normalized problems while still mixing heterogeneous channels at each timestep. In short, the former preserves temporal completeness but mixes action-channel geometry, whereas the latter reduces group dimensionality but breaks each channel's temporal trajectory. These limitations motivate the remaining natural orientation: organizing drifting along the complete temporal trajectory of each individual action channel.

To address these two challenges, we introduce DriftingVLA, a native one-step VLA that combines pretrained multimodal representations with direct distribution-drifting action generation. Figure~\ref{fig:overview} summarizes the shift from iterative robot-time refinement to training-time distribution drifting and direct one-step deployment. For the generative interface, DriftingVLA retains the pretrained multimodal backbone, freshly initializes the continuous action-generation branch, and jointly post-trains both components so that the action expert learns a direct mapping from action-shaped Gaussian noise to a complete action chunk rather than inheriting a flow-velocity parameterization. For the structured action geometry, we propose Per-Dimension Temporal Drifting (PDTD), which treats each channel trajectory $\mathbf A_{:,d}\in\mathbb R^H$ as one drifting group and defines its distributional geometry independently of heterogeneous action dimensions. Crucially, PDTD factorizes only the geometry used to construct the training signal, not the action generator itself. The shared VLM and action Transformer still jointly generate the complete action chunk and retain the capacity to model cross-dimensional dependencies required for coordinated control. During training, the multimodal prefix is additionally shared across sibling samples so that distribution learning does not require repeated VLM computation.

We evaluate DriftingVLA on LIBERO, RoboTwin~2.0, and six real-world manipulation tasks spanning both single- and dual-arm control. On LIBERO, DriftingVLA achieves a $98.32\%$ overall success rate, compared with $97.10\%$ for the original 10-NFE $\pi_{0.5}$ policy and $97.48\%$ for SnapFlow. On RoboTwin~2.0, it reaches $81.09\%$, exceeding 10-NFE $\pi_{0.5}$ by 1.51 percentage points and SnapFlow by 2.56 points, while a naive one-step evaluation of the conventionally trained $\pi_{0.5}$ flow policy drops to $63.76\%$. In real-world evaluation, DriftingVLA reaches $77.67\%$ overall success, compared with $74.22\%$ for $\pi_{0.5}$, with improvements on both the single- and dual-arm subsets. The geometry ablation further shows that PDTD consistently improves over both Chunk-wise and Step-wise drifting on LIBERO and RoboTwin~2.0. Finally, under controlled A800 profiling, native one-step generation reduces action-chunk latency from $227.61$\,ms to $67.67$\,ms, corresponding to a $3.36\times$ speedup; increasing the number of drifting siblings from $G=2$ to $G=8$ raises measured training cost by only $3.78\%$ and peak memory by $9.54\%$. Together, these results show that iterative action refinement can be removed from deployment without sacrificing control performance.

The present article substantially extends the DBP formulation introduced in DBPO~\cite{gao2026drift} to pretrained VLAs. The resulting extensions and contributions are threefold:
\begin{itemize}

    \item We extend native drift-based generation to pretrained flow-based VLAs by preserving and jointly adapting the pretrained multimodal backbone while relearning the flow-specific action interface as a direct conditional generator, enabling one-step deployment without numerical integration or flow-trajectory distillation.

    \item We identify action-chunk organization as a key design variable when extending distribution drifting to VLAs and propose Per-Dimension Temporal Drifting (PDTD). In contrast to the Chunk-wise and Step-wise organizations considered in DBP, PDTD preserves complete per-channel temporal trajectories while separating heterogeneous action-channel geometries, without factorizing the joint action generator. 
    
    \item We provide an expanded evaluation across LIBERO, RoboTwin~2.0, and six real-world single- and dual-arm tasks, together with controlled studies of VLA adaptation, drifting geometry, and deployment efficiency. DriftingVLA outperforms representative one-step VLA approaches and multi-step $\pi_{0.5}$ while substantially reducing action-generation latency.
    
\end{itemize}
\section{Related Work}
\label{sec:related_work}

\subsection{Vision-Language-Action Models}
\label{sec:related_vla}

The development of vision-language-action (VLA) models builds on advances in large-scale language-conditioned robot learning, cross-embodiment data, and pretrained multimodal representations~\cite{palme2023,rt1_2023,openx2024}. RT-2~\cite{rt2_2023} demonstrated how vision-language models can be co-fine-tuned on web-scale data and robotic trajectories for robotic control, while Octo~\cite{octo2024} and OpenVLA~\cite{openvla2025} further developed large-scale, reusable robot policies across diverse tasks and embodiments. A central design choice in these models is the action interface~\cite{openvlaoft2025,rdt1b2024,cogact2024,grootn1_2025,arvla2026}. FAST~\cite{fast_2025} compresses action sequences into frequency-space tokens for efficient autoregressive control, whereas $\pi_0$~\cite{pi0_2024} and $\pi_{0.5}$~\cite{pi05_2024} pair a pretrained multimodal backbone with a flow-matching action expert for continuous action-chunk generation. These developments highlight a useful separation between multimodal representation learning and action generation. DriftingVLA focuses on the latter, retaining the pretrained multimodal representation while redesigning the continuous action generator for direct one-step control.

\subsection{Efficient and One-Step Generative Action Policies}
\label{sec:related_onestep}

The inference cost of iterative generative policies has motivated several forms of acceleration~\cite{frans2024one,meanflow2024,asyncvla2025,falcon2025,reactvla2026,onestepflowpolicy2026}. For diffusion-based visuomotor policies, Consistency Policy~\cite{prasad2024consistency} and One-Step Diffusion Policy~\cite{wang2024onestepdiffusionpolicyfast} distill pretrained diffusion policies into few- or one-step generators. For flow-based VLAs, execution-level approaches instead retain the iterative generator while reducing its impact on online control. Real-Time Chunking (RTC)~\cite{rtc_2025} overlaps action generation with execution, while FASTER~\cite{faster_2026} prioritizes near-term actions through horizon-aware sampling. These approaches improve responsiveness without replacing the underlying iterative generative process.

Other methods directly target one-step action generation through changes to the training or action-generation formulation~\cite{yan2025maniflow,sheng2025mp1,Fang2025OMPOM,letitbesimple2026,invertibleadapter2026,mmact2025,onestepflowpolicy2026}. SnapFlow~\cite{snapflow2026} uses progressive self-distillation to compress multi-step flow sampling into 1-NFE generation, while MeanFlowVLA~\cite{meanflowvla_2026} adapts the MeanFlow principle~\cite{meanflow2024} to predict interval-averaged velocity for one-step action generation. $\pi_0$-EqM~\cite{pi0eqm_2026} explores an alternative equilibrium-matching decoder with adaptive inference depth. DriftingVLA differs in how one-step behavior is obtained: instead of deriving the deployed generator from an inference-time transport process, it trains a direct conditional generator whose output distribution is refined during optimization.

\subsection{Drifting Models and One-Step Robot Policies}
\label{sec:related_drifting}

Drifting Models~\cite{deng2026generative} formulate generative learning through the pushforward distribution of a generator, using distribution-level corrections during training instead of iterative denoising or transport at inference. Drift-Based Policy Optimization (DBPO)~\cite{gao2026drift} introduced a native one-step policy learning framework for robotic control, whose generative component, Drift-Based Policy (DBP), applies distribution drifting in action space using multiple sibling action hypotheses during training while retaining single-sample, one-evaluation deployment.

DriftingVLA builds on the DBP generative formulation and studies the additional issues that arise when distribution drifting is extended to a pretrained flow-based VLA. First, the flow-specific action interface must be converted into a direct generator without discarding the pretrained multimodal representation. Second, the structured $H\times D$ action chunk requires an explicit choice of distributional geometry. Whereas DBP considers Chunk-wise and Step-wise organizations for action-chunk drifting, DriftingVLA introduces Per-Dimension Temporal Drifting (PDTD), which defines channel-specific geometry over complete temporal trajectories while retaining joint action generation through the shared VLA model.
\section{Preliminaries}\label{sec:preliminaries}

We first review the flow-based action generation mechanism used by $\pi_{0.5}$ and the drift-based policy learning framework introduced in prior work. Throughout the paper, we denote the conditioning context at control step $t$ by
$
\mathbf{c}_t=(\mathbf{I}_t^{1:N_v},\ell_t,\mathbf{s}_t),
$
which consists of multi-view visual observations $\mathbf{I}_t^{1:N_v}$, a language instruction $\ell_t$, and the robot proprioceptive state $\mathbf{s}_t$. The policy predicts an action chunk
\begin{equation}
\mathbf{A}_t
=
[\mathbf{a}_t^{1},\ldots,\mathbf{a}_t^{H}]^{\top}
\in \mathbb{R}^{H\times D},
\label{eq:action_chunk}
\end{equation}
where $\mathbf{a}_t^h\in\mathbb{R}^{D}$ denotes the action at future step $h$, $H$ denotes the prediction horizon, and $D$ denotes the per-step action dimension. We use $h\in\{1,\ldots,H\}$ and $d\in\{1,\ldots,D\}$ to index the temporal and action-dimension axes, respectively.

\subsection{Flow-Based Action Generation in $\pi_{0.5}$}\label{sec:prelim_flow}

$\pi_{0.5}$~\cite{pi05_2024} builds on the flow-based VLA architecture introduced by $\pi_0$~\cite{pi0_2024}. Its low-level continuous action generator consists of a pretrained vision-language backbone together with a smaller action expert specialized for processing robot states and continuous action tokens. During post-training, the multimodal context forms a shared prefix, while the action expert receives noisy action tokens and predicts the corresponding flow-matching vector field.

Specifically, conditional flow matching~\cite{lipman2023flow,rectifiedflow2022} constructs an interpolation between Gaussian noise and the demonstrated action chunk. Given a demonstration $\mathbf{A}_t$ and Gaussian noise
$
\boldsymbol{\epsilon}\sim\mathcal{N}(\mathbf{0},\mathbf{I}),
$
the noisy action at flow time $\tau\in[0,1]$ is
\begin{equation}
\mathbf{A}_t^{\tau}
=
\tau\mathbf{A}_t
+
(1-\tau)\boldsymbol{\epsilon}.
\label{eq:flow_interpolation}
\end{equation}
The action expert parameterizes a conditional velocity field
$
\mathbf{v}_{\boldsymbol{\theta}}
(\mathbf{A}_t^{\tau},\mathbf{c}_t,\tau)
$
and is trained to match the target vector field
$
\mathbf{A}_t-\boldsymbol{\epsilon}
$:
\begin{equation}
\mathcal{L}_{\mathrm{FM}}
=
\mathbb{E}_{\mathbf{A}_t,\mathbf{c}_t,\tau,\boldsymbol{\epsilon}}
\left[
\left\|
\mathbf{v}_{\boldsymbol{\theta}}
(\mathbf{A}_t^{\tau},\mathbf{c}_t,\tau)
-
(\mathbf{A}_t-\boldsymbol{\epsilon})
\right\|_{F}^{2}
\right].
\label{eq:flow_matching_loss}
\end{equation}
Here, $\boldsymbol{\theta}$ denotes the model parameters and $\|\cdot\|_F$ denotes the Frobenius norm over the action chunk.

At inference time, action generation starts from
$
\mathbf{A}_t^{0}\sim\mathcal{N}(\mathbf{0},\mathbf{I})
$
and numerically integrates the learned vector field from $\tau=0$ to $\tau=1$. Using forward Euler integration with $K$ steps and step size $\delta=1/K$,
\begin{equation}
\mathbf{A}_t^{\tau_{k+1}}
=
\mathbf{A}_t^{\tau_k}
+
\delta\,
\mathbf{v}_{\boldsymbol{\theta}}
(\mathbf{A}_t^{\tau_k},\mathbf{c}_t,\tau_k),
\qquad
\tau_k=\frac{k}{K}.
\label{eq:flow_euler}
\end{equation}
The final action chunk is obtained as $\widehat{\mathbf{A}}_t=\mathbf{A}_t^{1}$. Following $\pi_0$, the multimodal prefix can be cached across integration steps, such that repeated computation is concentrated in the action expert. Nevertheless, generating one action chunk still requires $K$ successive evaluations of the action expert, motivating action-generation mechanisms that avoid iterative inference altogether.

\subsection{Drift-Based Policy Learning}\label{sec:prelim_dbp}

Drifting Models (DM)~\cite{deng2026generative} provide a different generative principle in which distribution refinement occurs through optimization rather than through an iterative inference trajectory. Drift-Based Policy (DBP), introduced as the generative component of Drift-Based Policy Optimization (DBPO)~\cite{gao2026drift}, specializes this principle to robotic control and enables direct one-step action generation.

We first summarize the underlying drifting formulation. Let $\mathbf{z}\sim p_0$ denote a latent sample from a fixed Gaussian prior and let
$
\mathbf{x}=f_{\boldsymbol{\theta}}(\mathbf{z})
\in\mathbb{R}^{S}
$
denote a generic generated sample, where $S$ is the dimensionality of the drifting unit. The generator induces the pushforward distribution
\begin{equation}
q_{\boldsymbol{\theta}}
=
[f_{\boldsymbol{\theta}}]{}_{\#}p_0.
\label{eq:drift_pushforward}
\end{equation}
Let $p_{\mathrm{data}}$ denote the target data distribution. At optimization iteration $k$, the current parameters induce a distribution $q_k$. For a fixed latent seed, DM conceptually evolves the generated sample according to
\begin{equation}
\mathbf{x}_{k+1}
=
\mathbf{x}_k
+
\mathcal{V}_{p_{\mathrm{data}},q_k}(\mathbf{x}_k),
\label{eq:drift_update}
\end{equation}
where
$
\mathcal{V}_{p,q}(\mathbf{x})
$
is a drifting field that specifies the distribution-level correction at $\mathbf{x}$. DM employs an anti-symmetric construction satisfying
$
\mathcal{V}_{p,q}(\mathbf{x})
=
-\mathcal{V}_{q,p}(\mathbf{x}),
$
and consequently
$
\mathcal{V}_{p,p}(\mathbf{x})=\mathbf{0}
$
at distributional equilibrium.

Rather than explicitly applying Eq.~\eqref{eq:drift_update} during inference, the correction is converted into a stop-gradient regression target during training:
\begin{equation}
\widetilde{\mathbf{x}}
=
\operatorname{sg}
\left(
f_{\boldsymbol{\theta}}(\mathbf{z})
+
\mathcal{V}_{p_{\mathrm{data}},q_{\boldsymbol{\theta}}}
\left(
f_{\boldsymbol{\theta}}(\mathbf{z})
\right)
\right),
\label{eq:drift_target}
\end{equation}
where $\operatorname{sg}(\cdot)$ denotes the stop-gradient operation. The generator is optimized by
\begin{equation}
\mathcal{L}_{\mathrm{DM}}
=
\mathbb{E}_{\mathbf{z}\sim p_0}
\left[
\left\|
f_{\boldsymbol{\theta}}(\mathbf{z})
-
\widetilde{\mathbf{x}}
\right\|_2^2
\right].
\label{eq:drift_loss}
\end{equation}
Repeated optimization progressively absorbs these distributional corrections into the generator parameters. Consequently, once training is complete, generation requires only the direct mapping $f_{\boldsymbol{\theta}}(\mathbf{z})$, without performing the drifting updates in Eq.~\eqref{eq:drift_update} at inference time.

In practice, the drifting field is estimated through interactions between target and generated samples. A generic kernelized form is
\begin{equation}
\mathcal{V}_{p,q}(\mathbf{x})
=
\mathbb{E}_{\mathbf{y}^{+}\sim p,\,
\mathbf{y}^{-}\sim q}
\left[
\mathcal{K}
(\mathbf{x},\mathbf{y}^{+},\mathbf{y}^{-})
\right],
\label{eq:kernel_drift}
\end{equation}
where $\mathbf{y}^{+}$ and $\mathbf{y}^{-}$ denote target and generated references, respectively. The resulting interaction combines attraction toward target samples with repulsion from generated samples while preserving the anti-symmetric drifting construction.

DBP applies this principle conditionally in robot action space. Given a training pair $(\mathbf{c}_i,\mathbf{A}_i)$, it draws $G$ independent latent samples and generates multiple sibling hypotheses,
\begin{equation}
\mathbf{Z}_i^{(g)}
\overset{\mathrm{i.i.d.}}{\sim} p_0,
\qquad
\widehat{\mathbf{A}}_i^{(g)}
=
f_{\boldsymbol{\theta}}
(\mathbf{Z}_i^{(g)},\mathbf{c}_i),
\quad
g=1,\ldots,G,
\label{eq:dbp_siblings}
\end{equation}
which provide samples from the conditional generated distribution
\begin{equation}
q_{\boldsymbol{\theta}}
(\cdot\mid\mathbf{c}_i)
=
\left[
f_{\boldsymbol{\theta}}
(\cdot,\mathbf{c}_i)
\right]{}_{\#}p_0.
\label{eq:conditional_pushforward}
\end{equation}
The drifting objective then shapes these hypotheses using demonstrated action chunks as positive references and generated hypotheses as negative references. Importantly, the multiple siblings are required only during training. At deployment, a single latent sample $\mathbf{Z}_t$ is drawn and a complete action chunk is obtained through one evaluation,
\begin{equation}
\widehat{\mathbf{A}}_t
=
f_{\boldsymbol{\theta}}
(\mathbf{Z}_t,\mathbf{c}_t).
\label{eq:dbp_inference}
\end{equation}

DBP considers two ways of organizing an action chunk for drifting. In Chunk-wise Drifting, the complete action chunk is flattened and treated as a single drifting unit,
\begin{equation}
\mathbf{x}_{\mathrm{chunk}}
=
\operatorname{vec}(\mathbf{A}_t)
\in\mathbb{R}^{HD}.
\label{eq:chunk_drifting}
\end{equation}
In Step-wise Drifting, the drifting objective is instead applied independently to each temporal slice,
\begin{equation}
\mathbf{x}_{\mathrm{step}}^{h}
=
\mathbf{A}_{t,h,:}
\in\mathbb{R}^{D},
\qquad h=1,\ldots,H,
\label{eq:stepwise_drifting}
\end{equation}
and the resulting losses are averaged across the prediction horizon. These two organizations define the action-chunk groupings considered in DBP. In the next section, we revisit this design when extending DBP to pretrained VLA models.

\section{Method}\label{sec:method}

\begin{figure*}[t]
    \centering
    \includegraphics[width=0.99\textwidth]{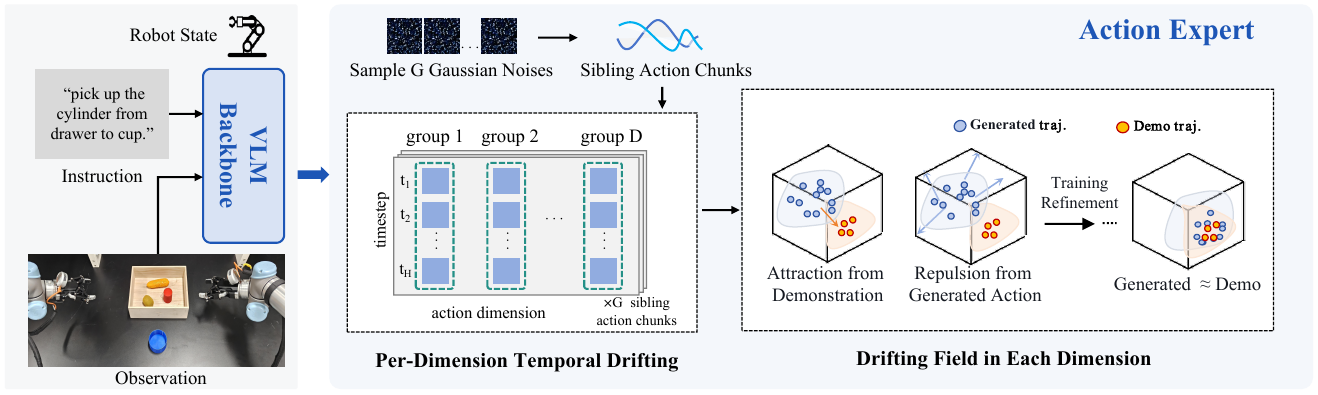}
    \caption{Training pipeline of DriftingVLA. The pretrained \(\pi_{0.5}\) multimodal backbone encodes each condition once, and the resulting shared prefix is reused across \(G\) sibling generations from independent Gaussian latent samples. A freshly initialized one-step action expert generates sibling action chunks in parallel, which are reorganized by PDTD into per-dimension temporal trajectories. The drifting field constructs attraction--repulsion targets from generated siblings and demonstrations, and the resulting detached targets jointly post-train the VLM backbone and action expert.}
    \label{fig:method_overview}
\end{figure*}
We introduce DriftingVLA, a VLA policy with native one-step action generation. Extending the DBP formulation in Sec.~\ref{sec:prelim_dbp} to a pretrained flow-based VLA requires addressing two coupled design problems: adapting the flow-specific action interface to direct generation while retaining pretrained multimodal representations, and defining an appropriate drifting geometry for the structured action chunk. DriftingVLA addresses the first by reconfiguring the $\pi_{0.5}$ action-generation branch as a direct noise-to-action generator and jointly post-training it with the pretrained VLM, and the second through Per-Dimension Temporal Drifting (PDTD), which organizes distribution drifting along each action channel's complete temporal trajectory.

\subsection{From Flow-Based VLA Models to Direct Action Generation in DriftingVLA}\label{sec:method_formulation}

\paragraph{Direct conditional generation}
At control step $t$, DriftingVLA generates a complete action chunk $\widehat{\mathbf{A}}_t\in\mathbb{R}^{H\times D}$ from the multimodal context $\mathbf{c}_t$. Instead of learning the time-dependent velocity field used by $\pi_{0.5}$ and recovering the final action through numerical integration, we directly parameterize the deployed policy as
\begin{equation}
\widehat{\mathbf{A}}_t
=
f_{\boldsymbol{\theta}}
(\mathbf{Z}_t,\mathbf{c}_t),
\qquad
\mathbf{Z}_t\sim\mathcal{N}(\mathbf{0},\mathbf{I}),
\label{eq:driftingvla_generator}
\end{equation}
where $\mathbf{Z}_t\in\mathbb{R}^{H\times D}$ is an action-shaped latent sampled from the Gaussian prior $p_0$ introduced in Sec.~\ref{sec:prelim_dbp}. The output of $f_{\boldsymbol{\theta}}$ is interpreted directly as the final action chunk rather than as a local update along a generative trajectory. Hence, the mapping learned during training is also the mapping executed at deployment.

This distinction separates DriftingVLA from evaluating a conventionally trained flow policy with only one Euler step. The latter changes the numerical solver while retaining a velocity estimator trained for local transport. DriftingVLA changes the learned function itself: the action expert is optimized from the outset to map latent noise directly to an action chunk. Its one-step behavior is therefore a property of the training formulation, not an inference-time approximation.

\paragraph{Preserving the pretrained VLA representation}
We retain the two-component $\pi_{0.5}$ architecture and write
\begin{equation}
\boldsymbol{\theta}
=
(\boldsymbol{\theta}_{\mathrm{VLM}},
 \boldsymbol{\theta}_{\mathrm{AE}}),
\label{eq:model_partition}
\end{equation}
where $\boldsymbol{\theta}_{\mathrm{VLM}}$ denotes the pretrained multimodal backbone and $\boldsymbol{\theta}_{\mathrm{AE}}$ denotes the continuous action-generation branch. We initialize $\boldsymbol{\theta}_{\mathrm{VLM}}$ from pretrained $\pi_{0.5}$, while freshly initializing the action expert together with its continuous-action input/output projections. During post-training, both parameter sets are optimized jointly.

This design preserves the pretrained multimodal representation while relearning the generative action interface. The pretrained VLM already provides vision-language representations suitable for conditioning robot behavior, whereas the original action expert is specialized for estimating a flow vector field. Retaining the former while relearning the latter avoids discarding the pretrained semantic and perceptual priors, yet does not constrain the new generator to inherit the previous velocity-field parameterization.

\paragraph{Shared-context sibling generation}
Distribution drifting requires multiple samples from the current conditional generator during training. For each training pair $(\mathbf{c}_i,\mathbf{A}_i)$, we draw $G$ independent latent chunks as in Eq.~\eqref{eq:dbp_siblings}. Since all siblings share the same multimodal condition, we factor generation into a shared context encoder and sibling-specific action generation:
\begin{equation}
\begin{aligned}
\mathbf{P}_i
&=E_{\boldsymbol{\theta}_{\mathrm{VLM}}}(\mathbf{c}_i),\\
\widehat{\mathbf{A}}_i^{(g)}
&=F_{\boldsymbol{\theta}_{\mathrm{AE}}}
(\mathbf{Z}_i^{(g)};\mathbf{P}_i),
\quad g=1,\ldots,G.
\end{aligned}
\label{eq:shared_sibling_generation}
\end{equation}
Here, $E_{\boldsymbol{\theta}_{\mathrm{VLM}}}$ denotes the multimodal context encoder and $F_{\boldsymbol{\theta}_{\mathrm{AE}}}$ denotes the action-generation branch conditioned on the shared context. The context is reused across all $G$ siblings, avoiding repeated multimodal computation, while gradients from all sibling losses jointly update the pretrained VLM during post-training. Because the siblings differ only in their independently sampled action latents, they remain samples from the same conditional generator $q_{\boldsymbol{\theta}}(\cdot\mid\mathbf{c}_i)$ defined in Sec.~\ref{sec:prelim_dbp}. Thus, shared-context sampling reduces the cost of conditional distribution learning without changing the conditional distribution being optimized.

Figure~\ref{fig:method_overview} summarizes the complete DriftingVLA training pipeline.

The formulation above specifies how drifting is integrated into a pretrained VLA, while the organization of the $H\times D$ action chunk remains an important design choice in the drifting objective. This grouping determines which coordinates share pairwise distances, scale statistics, affinities, and normalized drifting forces.

\subsection{From Action-Chunk Geometry to Per-Dimension Temporal Drifting}\label{sec:method_geometry}

An action chunk $\mathbf{A}\in\mathbb{R}^{H\times D}$ has two structurally different axes~\cite{act2023,diffusionpolicy2023,openvlaoft2025,baku2024,fast_2025,dp3_2024}. The temporal slice $\mathbf{A}_{h,:}\in\mathbb{R}^{D}$ collects heterogeneous control variables at one future step, whereas the column $\mathbf{A}_{:,d}\in\mathbb{R}^{H}$ describes the complete temporal evolution of one action channel. This distinction matters for drifting because distances determine the neighborhood structure used to construct attraction and repulsion updates.

The two organizations reviewed in Sec.~\ref{sec:prelim_dbp} induce different geometries. Chunk-wise drifting uses the entire flattened chunk as one $HD$-dimensional unit. For two chunks $\mathbf{A}$ and $\mathbf{B}$,
\begin{equation}
\|\operatorname{vec}(\mathbf{A})-
  \operatorname{vec}(\mathbf{B})\|_2^2
=
\sum_{d=1}^{D}\sum_{h=1}^{H}
(A_{h,d}-B_{h,d}){}^2.
\label{eq:chunk_distance}
\end{equation}
It therefore preserves the complete horizon, but all action channels jointly determine a single distance, a single scale, and the same distribution-level neighborhood statistics. A discrepancy in one channel changes the geometry that governs the drifting signal applied to every other channel. We refer to this effect as cross-channel geometric interference. This does not imply that cross-channel coupling in the policy is undesirable; coordinated robot control clearly requires it. The issue is specifically whether heterogeneous control channels must also share one distributional metric and normalization.

Step-wise drifting reduces the scope of each group by treating $\mathbf{A}_{h,:}$ independently for $h=1,\ldots,H$. This allows different future timesteps to have separate drifting statistics, but fragments the complete trajectory $\mathbf{A}_{:,d}=[A_{1,d},\ldots,A_{H,d}]{}^\top$ across $H$ independently normalized problems. Consequently, the drifting objective no longer directly compares two samples according to the full temporal evolution of channel $d$. Moreover, all $D$ heterogeneous channels at a given timestep still jointly determine the same local distance.

These limitations motivate organizing the action tensor along its remaining natural orientation. We introduce Per-Dimension Temporal Drifting (PDTD), which treats the full temporal trajectory of each action dimension as one drifting unit:
\begin{equation}
\mathcal{G}_{\mathrm{PDTD}}(\mathbf{A})
=
\{\mathbf{A}_{:,d}\}{}_{d=1}^{D},
\qquad
\mathbf{A}_{:,d}\in\mathbb{R}^{H}.
\label{eq:pdtd_grouping}
\end{equation}
Hence, Chunk-wise, Step-wise, and PDTD respectively define drifting units of dimensions $HD$, $D$, and $H$. More importantly, they assign the temporal and action-channel axes different roles: Chunk-wise collapses both axes into one geometry; Step-wise indexes groups by time and mixes channels inside each group; PDTD indexes groups by action channel and preserves the complete temporal horizon inside each group.

\begin{figure}[!t]
    \centering
    \includegraphics[width=\columnwidth]{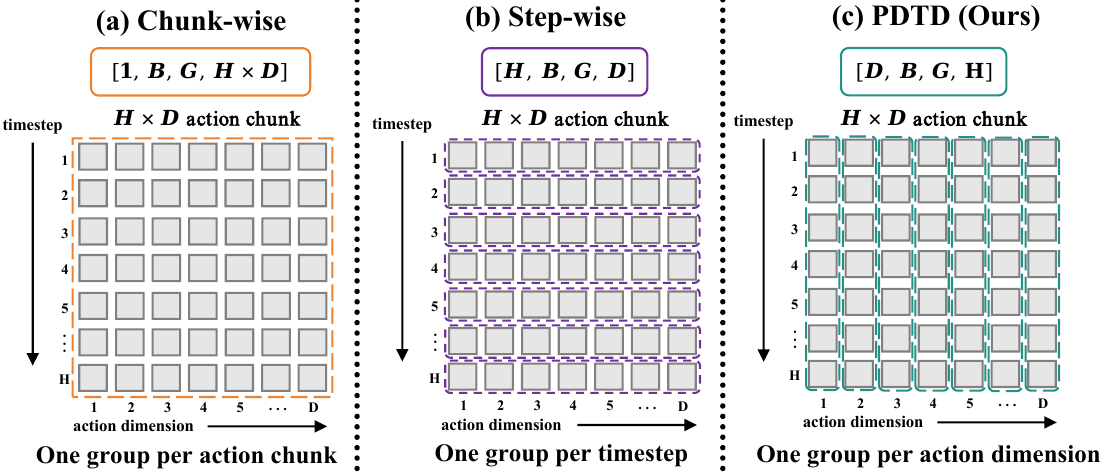}
    \caption{Action-chunk grouping strategies for distribution drifting. Chunk-wise drifting treats an \(H\times D\) action chunk as one \(HD\)-dimensional group; Step-wise drifting forms \(H\) groups of dimension \(D\); and PDTD forms \(D\) groups of dimension \(H\). PDTD preserves the complete temporal trajectory of each action channel while defining its drifting geometry separately. The grouping affects only the drifting geometry; the full action chunk is still generated jointly by the shared VLA model.}
    \label{fig:drifting_grouping}
\end{figure}

Figure~\ref{fig:drifting_grouping} visualizes the three organizations and their corresponding group shapes.

For sibling $g$ of training example $i$, PDTD extracts
\begin{equation}
\mathbf{x}_{i,g}^{(d)}
=
\widehat{\mathbf{A}}_{i,:,d}^{(g)},
\qquad
\mathbf{x}_{i,+}^{(d)}
=
\mathbf{A}_{i,:,d},
\label{eq:pdtd_trajectories}
\end{equation}
so that, for every channel $d$, the $G$ siblings provide multiple predicted $H$-step trajectories under the same condition. Pairwise distances, data-dependent distance scales, affinities, and force normalizations are then estimated independently for each $d$. PDTD therefore gives each action channel its own temporal drifting geometry while retaining the complete trajectory as the primitive distributional object.

This organization is particularly suitable for VLA action interfaces whose channels correspond to heterogeneous controls such as translation, rotation, gripper commands, or other embodiment-dependent quantities. These channels can differ substantially in numerical range and in the temporal variations that characterize similar behavior. PDTD therefore allows each channel to maintain its own trajectory scale and similarity structure while still using the complete temporal trajectory as the primitive distributional object. The grouping imposes no particular temporal profile; it only determines how distributional similarity is measured when constructing the drifting signal.

\paragraph{Factorized geometry, joint action generation}
A crucial distinction is that PDTD factorizes the drifting geometry, not the action generator. The complete action chunk is generated jointly before any grouping is applied:
\begin{equation}
\widehat{\mathbf{A}}_i^{(g)}
=
f_{\boldsymbol{\theta}}
(\mathbf{Z}_i^{(g)},\mathbf{c}_i)
\in\mathbb{R}^{H\times D}.
\label{eq:pdtd_joint_generation}
\end{equation}
PDTD is applied only afterward to construct the training signal. Accordingly, we neither assume nor enforce
\begin{equation}
q_{\boldsymbol{\theta}}
(\mathbf{A}\mid\mathbf{c})
=
\prod_{d=1}^{D}
q_{\boldsymbol{\theta}}
(\mathbf{A}_{:,d}\mid\mathbf{c}).
\label{eq:pdtd_no_independence}
\end{equation}
All per-dimension drifting losses update the same joint VLA generator. Thus, the prediction of one channel is not restricted to depend only on the corresponding latent coordinates, and cross-dimensional dependencies can still be represented through the shared conditional generator. This distinction is important for coordinated multi-axis and multi-joint control, where different control channels must remain coupled at the policy level even though their drifting geometries are defined separately. In short, PDTD removes geometry-level coupling without removing model-level coupling.

\subsection{Per-Dimension Drifting Objective}\label{sec:method_drift_objective}

We next specify how the drifting field is constructed once PDTD has defined the groups. The formulation follows the attraction and repulsion mechanism reviewed in Sec.~\ref{sec:prelim_dbp}; here we emphasize the quantities that become channel specific under PDTD. For a minibatch of $B$ conditions, all trajectory statistics and reductions below are evaluated only over valid temporal coordinates, with padded entries excluded.

For each condition $\mathbf{c}_i$ and channel $d$, we form a reference set containing the $G$ generated sibling trajectories and the demonstrated trajectory,
\begin{equation}
\mathcal{Y}_{i}^{(d)}
=
\{\mathbf{x}_{i,1}^{(d)},\ldots,
  \mathbf{x}_{i,G}^{(d)},
  \mathbf{x}_{i,+}^{(d)}\}.
\label{eq:pdtd_reference_set}
\end{equation}
Let $\mathbf{y}_{i,u}^{(d)}$ denote its $u$-th element. PDTD first estimates a separate distance ruler for every action channel,
\begin{equation}
\begin{aligned}
s_d
&=\operatorname{Mean}_{i,g,u}
\left[
\left\|
\mathbf{x}_{i,g}^{(d)}-
\mathbf{y}_{i,u}^{(d)}
\right\|_2
\right],\\
\Delta_{i,g,u}^{(d)}
&=\frac{
\left\|
\mathbf{x}_{i,g}^{(d)}-
\mathbf{y}_{i,u}^{(d)}
\right\|_2
}{\max(s_d,\varepsilon)}.
\end{aligned}
\label{eq:pdtd_normalized_distance}
\end{equation}
Here and below, small positive constants are used where needed for numerical stability. The ruler $s_d$ is estimated from the current generated--reference relations for channel $d$, so the normalized distance reflects trajectory variation within that channel rather than the aggregate scale of the full action tensor. A channel with a different numerical range or trajectory variability therefore obtains a different geometric ruler rather than inheriting the scale of other action dimensions.

From the normalized distances, we construct multi-scale attraction and repulsion interactions for bandwidths $\rho\in\mathcal{R}$. Let $\mathcal{I}^{-}=\{1,\ldots,G\}$ denote generated references and $\mathcal{I}^{+}=\{G+1\}$ the demonstrated reference. Self-connections between a sibling and its identical generated reference are excluded by defining
\begin{equation}
\widetilde{\Delta}_{i,g,u}^{(d)}
=
\begin{cases}
+\infty, & u=g,\;u\in\mathcal{I}^{-},\\
\Delta_{i,g,u}^{(d)}, & \text{otherwise},
\end{cases}
\qquad
\ell_{i,g,u}^{(\rho,d)}
=-\frac{\widetilde{\Delta}_{i,g,u}^{(d)}}{\rho}.
\label{eq:pdtd_affinity_logits}
\end{equation}
For each generated trajectory, one Softmax normalizes over candidate references and the other over competing generated siblings for a fixed reference:
\begin{equation}
\begin{aligned}
P_{i,g,u}^{\rightarrow(\rho,d)}
&=
\frac{\exp(\ell_{i,g,u}^{(\rho,d)})}
{\sum_{u'}\exp(\ell_{i,g,u'}^{(\rho,d)})},\\
P_{i,g,u}^{\leftarrow(\rho,d)}
&=
\frac{\exp(\ell_{i,g,u}^{(\rho,d)})}
{\sum_{g'}\exp(\ell_{i,g',u}^{(\rho,d)})},\\
W_{i,g,u}^{(\rho,d)}
&=
\sqrt{
\max\!\left(
P_{i,g,u}^{\rightarrow(\rho,d)}
P_{i,g,u}^{\leftarrow(\rho,d)},
\varepsilon_a
\right)} .
\end{aligned}
\label{eq:pdtd_softmax_affinity}
\end{equation}
The geometric mean yields a bidirectionally normalized affinity, assigning large weight only when the generated trajectory and the reference mutually favor the interaction. Applying this construction independently for every $d$ gives each action channel its own neighborhood structure, while using multiple bandwidths allows the field to aggregate interactions at different trajectory scales.

We next balance attraction to demonstrations and repulsion among generated siblings. Define the negative and positive affinity masses
\begin{equation}
S_{i,g,-}^{(\rho,d)}
=
\sum_{u\in\mathcal{I}^{-}}
W_{i,g,u}^{(\rho,d)},
\qquad
S_{i,g,+}^{(\rho,d)}
=
\sum_{u\in\mathcal{I}^{+}}
W_{i,g,u}^{(\rho,d)} .
\label{eq:pdtd_affinity_mass}
\end{equation}
The signed coefficients are
\begin{equation}
\alpha_{i,g,u}^{(\rho,d)}
=
\begin{cases}
-\,W_{i,g,u}^{(\rho,d)}
S_{i,g,+}^{(\rho,d)},
& u\in\mathcal{I}^{-},\\[1mm]
\phantom{-}\,W_{i,g,u}^{(\rho,d)}
S_{i,g,-}^{(\rho,d)},
& u\in\mathcal{I}^{+}.
\end{cases}
\label{eq:pdtd_alpha}
\end{equation}
Thus generated references contribute repulsive coefficients whereas the demonstrated reference contributes an attractive coefficient. The cross-side mass coupling gives
\begin{equation}
\sum_{u\in\mathcal{I}^{-}}
\alpha_{i,g,u}^{(\rho,d)}
=
-
\sum_{u\in\mathcal{I}^{+}}
\alpha_{i,g,u}^{(\rho,d)},
\qquad
\sum_u \alpha_{i,g,u}^{(\rho,d)}=0,
\label{eq:pdtd_mass_balance}
\end{equation}
so the field is translation invariant and can be expressed through relative displacements. This balance also prevents the net coefficient mass from being dominated by one side of the attraction--repulsion interaction.

For force construction, we normalize trajectory coordinates using
\begin{equation}
\eta_d=\max(s_d/\sqrt{H},\varepsilon),
\qquad
\overline{\mathbf{x}}_{i,g}^{(d)}
=\frac{\mathbf{x}_{i,g}^{(d)}}{\eta_d},
\qquad
\overline{\mathbf{y}}_{i,u}^{(d)}
=\frac{\mathbf{y}_{i,u}^{(d)}}{\eta_d}.
\label{eq:pdtd_coordinate_normalization}
\end{equation}
The scale-specific force and the resulting multi-scale drifting field are
\begin{equation}
\begin{aligned}
\mathbf{F}_{i,g}^{(\rho,d)}
&=
\sum_{u=1}^{G+1}
\alpha_{i,g,u}^{(\rho,d)}
\left(
\overline{\mathbf{y}}_{i,u}^{(d)}-
\overline{\mathbf{x}}_{i,g}^{(d)}
\right),\\
r_{\rho,d}
&=
\operatorname{RMS}_{i,g}
\left(
\mathbf{F}_{i,g}^{(\rho,d)}
\right),\\
\mathbf{V}_{i,g}^{(d)}
&=
\sum_{\rho\in\mathcal{R}}
\frac{\mathbf{F}_{i,g}^{(\rho,d)}}{\max(r_{\rho,d},\varepsilon_f)}.
\end{aligned}
\label{eq:pdtd_multiscale_field}
\end{equation}
Here, the RMS is evaluated over the minibatch, sibling samples, and valid temporal coordinates. The multi-bandwidth construction combines local and broader neighborhood information, while the per-$(\rho,d)$ normalization prevents a particular scale or action channel from dominating because of its raw numerical magnitude. Together with the channel-specific ruler $s_d$, this normalization carries the per-dimension factorization through the complete drifting-field construction rather than applying it only at the initial grouping step. Thus, PDTD makes the complete sequence of geometric statistics---distance scale, affinity, attraction--repulsion coupling, and force normalization---channel specific.

As in DBP, the field is converted into a frozen regression target rather than differentiated through:
\begin{equation}
\widetilde{\mathbf{x}}_{i,g}^{(d)}
=
\operatorname{sg}
\left[
\overline{\mathbf{x}}_{i,g}^{(d)}+
\mathbf{V}_{i,g}^{(d)}
\right].
\label{eq:pdtd_target}
\end{equation}
The PDTD objective is the mean squared regression error over all valid trajectory coordinates,
\begin{equation}
\mathcal{L}_{\mathrm{PDTD}}
=
\operatorname{Mean}_{i,d,g,h\in\operatorname{valid}(i)}
\left[
\left(
\overline{x}_{i,g,h}^{(d)}-
\widetilde{x}_{i,g,h}^{(d)}
\right)^2
\right].
\label{eq:pdtd_loss}
\end{equation}
The stop-gradient target is recomputed from samples of the current generator at every optimization iteration. Each update therefore regresses the current sibling trajectories toward targets constructed from the current conditional distribution and the demonstrated trajectory. Repeated optimization absorbs these distribution-level corrections into the generator parameters, following the DBP principle reviewed in Sec.~\ref{sec:prelim_dbp}. Distribution refinement consequently occurs across parameter updates during training; the drifting field itself is not evaluated when the trained policy is deployed.

\subsection{Training and Native One-Step Inference}\label{sec:method_training_inference}

\paragraph{Padding-aware action space}
PDTD is evaluated only over valid physical action dimensions and valid temporal coordinates. When the underlying VLA uses a padded action representation to accommodate different embodiments or variable valid horizons, generated and demonstrated chunks are restricted to their valid coordinates before the PDTD geometry is constructed. Padded action dimensions and future timesteps therefore contribute neither to distance statistics nor to affinities or drifting forces.

\paragraph{Post-training}
During post-training, each condition produces $G$ sibling action chunks from independent latent samples. PDTD reorganizes these chunks into per-dimension temporal trajectories, constructs channel-specific drifting targets from the siblings and the demonstrated chunk, and applies Eq.~\eqref{eq:pdtd_loss} to update the policy. The resulting objective jointly adapts the pretrained VLM and the freshly initialized action-generation branch. The multimodal context is shared across siblings, so training remains multi-sample and distribution aware without repeating the condition encoding for each hypothesis.

\paragraph{Inference}
At deployment, all distribution-level machinery used above is removed. Given the current context $\mathbf{c}_t$, we draw one Gaussian latent chunk and evaluate Eq.~\eqref{eq:driftingvla_generator} once. No sibling set, pairwise interaction, drifting field, or numerical integration is required.

Consequently, DriftingVLA moves iterative refinement from robot-time inference to optimization-time distribution learning. The trained policy preserves the multimodal representation and joint action-generation capacity of the pretrained VLA, while deployment reduces to a single direct noise-to-action evaluation.

\section{Experiments}\label{sec:experiments}

We evaluate DriftingVLA from five complementary perspectives: (1) whether native one-step action generation can match or exceed iterative flow-based VLA control in simulation, (2) whether this performance transfers to real-world manipulation, (3) whether the proposed PDTD geometry improves over the Chunk-wise and Step-wise organizations inherited from DBP, (4) which components are important for adapting a pretrained VLA to the direct drifting objective, and (5) whether native one-step generation provides the intended deployment-time benefit without excessive training overhead.

\subsection{Experimental Setup}\label{sec:exp_setup}

\paragraph{Simulation benchmarks}
Our simulation evaluation uses LIBERO~\cite{libero2023} and RoboTwin~2.0~\cite{robotwin2025}. LIBERO contains four suites---Spatial, Object, Goal, and Long---with 10 tasks per suite. Each trained policy is evaluated for 100 episodes per task. RoboTwin~2.0 provides a substantially larger and more diverse bimanual manipulation benchmark. We evaluate all 50 tasks under both the Easy setting (\path{demo_clean}) and the domain-randomized Hard setting (\path{demo_randomized}), with 100 evaluation episodes for every task and difficulty level.

For both simulation benchmarks, every model is trained with three random seeds (42, 43, and 44) for 50k optimization steps. Unless otherwise stated, training uses 8 NVIDIA A800 80GB GPUs with a per-GPU batch size of 4. DriftingVLA uses $G=8$ generated siblings per condition. We report the mean success rate and sample standard deviation across the three seeds.

\begin{figure*}[t]
    \centering
    \includegraphics[width=\textwidth]{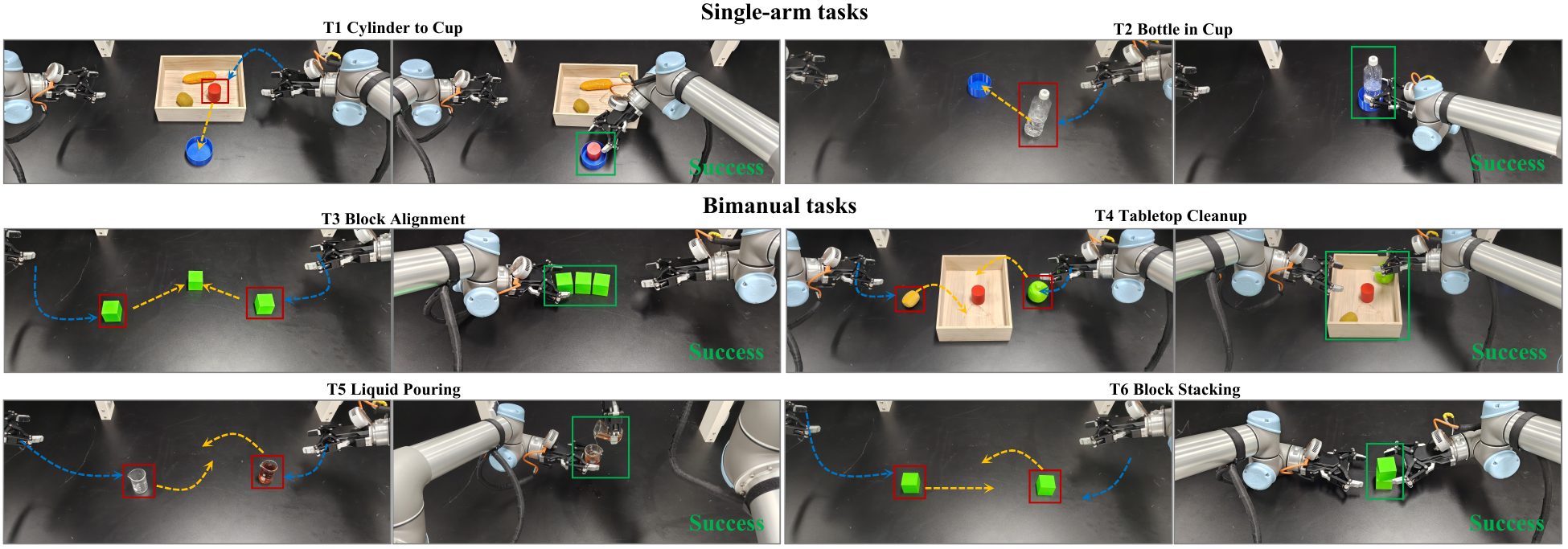}
    \caption{Real-world evaluation tasks. Initial and successful terminal states are shown for six manipulation tasks: two single-arm tasks (T1–T2) and four dual-arm tasks (T3–T6). The tasks are Cylinder to Cup, Bottle in Cup, Block Alignment, Tabletop Cleanup, Liquid Pouring, and Block Stacking.}
    \label{fig:realworld_tasks}
\end{figure*}

\paragraph{Real-world tasks}
We further evaluate on six real-world manipulation tasks, comprising two single-arm tasks and four dual-arm tasks. The platform consists of two UR5 manipulators, two wrist-mounted Intel RealSense L515 cameras, and one head-mounted Orbbec Gemini camera. Single-arm tasks use only the right arm, whereas dual-arm tasks require coordinated control of both manipulators. Following the task order used throughout the real-world evaluation, T1--T2 are Cylinder to Cup and Bottle in Cup, while T3--T6 are Block Alignment, Tabletop Cleanup, Liquid Pouring, and Block Stacking. Figure~\ref{fig:realworld_tasks} shows the initial and successful terminal states for all six tasks. We collect 100 teleoperated demonstrations for each task. Each method is trained for 50k optimization steps with three random seeds and evaluated for 50 trials per task and seed, yielding 300 evaluation trials per seed across the six-task suite. Robot-time inference runs on an NVIDIA RTX~3090.

\paragraph{Compared methods}
For the main simulation comparison, we use the original multi-step $\pi_{0.5}$~\cite{pi05_2024} with 10 action-expert network function evaluations (NFEs) per generated chunk, together with $\pi_{0.5}$-1step, obtained by evaluating the same conventionally trained flow policy with a single Euler step. We include the latter as a diagnostic solver-truncation control rather than as a natively trained one-step baseline. We also compare with two representative one-step VLA approaches: SnapFlow~\cite{snapflow2026}, which compresses flow-based generation through progressive self-distillation, and MeanFlowVLA~\cite{meanflowvla_2026}, which adapts the MeanFlow formulation to one-step VLA generation. We reimplement both one-step baselines following their published descriptions. For the geometry study, DriftingVLA-Chunk and DriftingVLA-Step replace PDTD with the Chunk-wise and Step-wise group organizations considered in DBP~\cite{gao2026drift}, respectively, while keeping the remaining DriftingVLA framework unchanged. All compared VLA policies use the same LeRobot $\pi_{0.5}$ base implementation and are trained under matched data and optimization budgets unless explicitly noted.

\subsection{Simulation Benchmark Results}\label{sec:exp_simulation}

Table~\ref{tab:simulation_main} reports the LIBERO comparison. DriftingVLA achieves an overall success rate of $98.32\%$, outperforming the 10-NFE $\pi_{0.5}$ policy by 1.22 percentage points and the strongest one-step baseline, SnapFlow, by 0.84 points. The margin is particularly informative on LIBERO-Long, where the benchmark is less saturated: DriftingVLA reaches $94.33\%$, compared with $92.40\%$ for $\pi_{0.5}$ and $93.07\%$ for SnapFlow. In contrast, the $\pi_{0.5}$-1step solver-truncation control reaches only $94.02\%$ overall. Thus, the one-step inference budget alone does not account for the performance of the direct generator.

\begin{table*}[!t]
\centering
\caption{LIBERO success rates (\%) across the Spatial, Object, Goal, and Long suites. Results are reported as mean \(\pm\) sample standard deviation over three training seeds, with 100 evaluation episodes per task. The best mean result in each column is shown in bold.}\label{tab:simulation_main}
\scriptsize
\renewcommand{\arraystretch}{1.08}
\setlength{\tabcolsep}{5.2pt}
\begin{tabular*}{0.94\textwidth}{@{\extracolsep{\fill}}lcccccc@{}}
\toprule
\multirow{2}{*}{Method} & \multirow{2}{*}{NFE} & \multicolumn{5}{c}{Success rate (\%)}\\
\cmidrule(lr){3-7}
& & Spatial & Object & Goal & Long & Overall\\
\midrule
$\pi_{0.5}$ & 10 & $98.40\pm0.20$ & $99.47\pm0.31$ & $98.13\pm0.12$ & $92.40\pm0.35$ & $97.10\pm0.13$\\
$\pi_{0.5}$-1step & 1 & $95.27\pm0.42$ & $96.00\pm0.20$ & $94.33\pm0.42$ & $90.47\pm0.31$ & $94.02\pm0.13$\\
\addlinespace[1.2pt]
SnapFlow & 1 & $99.20\pm0.53$ & $98.73\pm0.50$ & $98.93\pm0.23$ & $93.07\pm0.23$ & $97.48\pm0.16$\\
MeanFlowVLA & 1 & $94.87\pm0.42$ & $94.33\pm0.12$ & $93.20\pm0.40$ & $89.00\pm0.80$ & $92.85\pm0.35$\\
\addlinespace[1.2pt]
\textbf{DriftingVLA} & 1 & $\mathbf{99.73}\pm0.12$ & $\mathbf{99.87}\pm0.12$ & $\mathbf{99.33}\pm0.23$ & $\mathbf{94.33}\pm0.42$ & $\mathbf{98.32}\pm0.13$\\
\bottomrule
\end{tabular*}
\end{table*}

\begin{figure}[!t]
    \centering
    \includegraphics[width=\columnwidth]{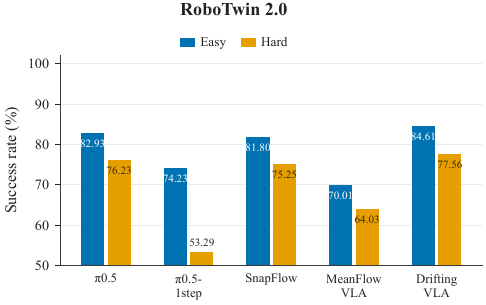}
    \caption{RoboTwin~2.0 success rates (\%) under the Easy and domain-randomized Hard settings. Results are averaged across all 50 tasks and three training seeds. \(\pi_{0.5}\) uses 10 NFEs, whereas \(\pi_{0.5}\)-1step, SnapFlow, MeanFlowVLA, and DriftingVLA perform one-step action generation.}
    \label{fig:robotwin_main}
\end{figure}

Figure~\ref{fig:robotwin_main} summarizes the RoboTwin~2.0 comparison, where the separation among methods becomes more pronounced. DriftingVLA reaches $81.09\%$ overall, exceeding 10-NFE $\pi_{0.5}$ by 1.51 points and SnapFlow by 2.56 points. On the harder domain-randomized split, DriftingVLA obtains $77.56\%$, compared with $76.23\%$ for $\pi_{0.5}$ and $75.25\%$ for SnapFlow. Under the same one-step solver truncation, $\pi_{0.5}$ drops to $63.76\%$ overall, including $53.29\%$ on the Hard split. DriftingVLA also substantially exceeds MeanFlowVLA on both simulation benchmarks. Overall, the main comparison shows that native one-step generation can retain or improve control performance relative to the original iterative policy and representative one-step VLA alternatives, while the large truncation gap motivates the controlled adaptation analysis in Sec.~\ref{sec:experiments_ablation}.

\subsection{Real-World Evaluation}\label{sec:exp_real}

Table~\ref{tab:realworld} reports the real-world results. DriftingVLA reaches the highest overall success rate at $77.67\%$, improving over 10-NFE $\pi_{0.5}$ ($74.22\%$) by 3.45 percentage points. Relative to the representative external one-step baselines, it improves over SnapFlow and MeanFlowVLA by 8.23 and 12.56 points, respectively. DriftingVLA achieves the best mean success rate on five of the six tasks; on T4 Tabletop Cleanup, $\pi_{0.5}$ is slightly higher ($76.67\%$ versus $75.33\%$).

The physical evaluation also includes the two controlled geometry variants inherited from DBP. DriftingVLA-Chunk reaches $67.78\%$ overall and DriftingVLA-Step reaches $66.11\%$, so PDTD improves over them by 9.89 and 11.56 points, respectively. The ordering $\text{Step}<\text{Chunk}<\text{PDTD}$ therefore persists from the LIBERO and RoboTwin~2.0 ablations to physical manipulation. This consistency provides additional evidence that action-chunk grouping is a consequential design choice rather than an implementation detail, while the complete simulation ablation in Sec.~\ref{sec:experiments_ablation} remains the controlled comparison of the three geometries.

DriftingVLA improves on both the single- and dual-arm subsets. Averaging the two single-arm tasks, success increases from $86.00\%$ for $\pi_{0.5}$ to $89.67\%$ for DriftingVLA, a 3.67-point gain. Across the four dual-arm tasks, the corresponding average increases from $68.34\%$ to $71.67\%$, a 3.33-point gain. The improvement on the dual-arm subset is relevant because these tasks require coordinated control across both manipulators. Although PDTD factorizes the drifting geometry across action dimensions, the full action chunk is still generated jointly by the shared VLM and Transformer action backbone, as described in Sec.~\ref{sec:method_geometry}. The real-world results therefore provide empirical evidence that per-dimension geometric supervision does not prevent coordinated multi-axis action generation.

\paragraph{Failure modes}
Observed failures include incorrect object localization, unstable grasps,
slippage, collisions, misalignment, and unstable placement. Dual-arm tasks
additionally exhibit arm-assignment, sequencing, inter-arm collision, and
coordination failures, while pouring tasks may fail because of inaccurate
alignment, insufficient transfer, or spillage. Trials exhibiting these
failure modes are counted as unsuccessful.

\begin{table*}[!t]
\centering
\caption{Real-world success rates (\%) across six manipulation tasks. T1–T2 are single-arm tasks and T3–T6 are bimanual tasks. Results are reported as mean \(\pm\) sample standard deviation over three training seeds, with 50 trials per task and seed. Overall aggregates all six tasks (300 trials per seed). The best mean result in each row is shown in bold.}\label{tab:realworld}
\scriptsize
\renewcommand{\arraystretch}{1.12}
\begin{tabular*}{0.99\textwidth}{@{\extracolsep{\fill}}>{\centering\arraybackslash}p{0.235\textwidth}cccccc@{}}
\toprule
\multirow{2}{*}{Task} &
\multicolumn{1}{c}{Reference} &
\multicolumn{2}{c}{One-step baselines} &
\multicolumn{2}{c}{Geometry variants} &
\multicolumn{1}{c}{Ours}\\
\cmidrule(lr){2-2}\cmidrule(lr){3-4}\cmidrule(lr){5-6}\cmidrule(lr){7-7}
& $\pi_{0.5}$ & SnapFlow & MeanFlowVLA & \makecell{DriftingVLA-\\Chunk} & \makecell{DriftingVLA-\\Step} & DriftingVLA\\
\midrule
\makecell[c]{\textbf{T1} Cylinder to Cup} &
$84.00\pm4.00$ & $74.67\pm3.06$ & $76.67\pm2.31$ & $76.67\pm1.15$ & $77.33\pm1.15$ & $\mathbf{87.33}\pm7.02$\\
\addlinespace[1.5pt]
\makecell[c]{\textbf{T2} Bottle in Cup} &
$88.00\pm4.00$ & $83.33\pm3.06$ & $82.00\pm3.46$ & $86.67\pm2.31$ & $84.67\pm1.15$ & $\mathbf{92.00}\pm2.00$\\
\midrule
\makecell[c]{\textbf{T3} Block Alignment} &
$62.67\pm5.03$ & $64.00\pm5.29$ & $52.67\pm8.08$ & $54.67\pm6.43$ & $52.67\pm8.33$ & $\mathbf{67.33}\pm10.07$\\
\addlinespace[1.5pt]
\makecell[c]{\textbf{T4} Tabletop Cleanup} &
$\mathbf{76.67}\pm6.43$ & $68.67\pm10.07$ & $66.00\pm11.14$ & $67.33\pm9.02$ & $66.67\pm7.02$ & $75.33\pm9.87$\\
\addlinespace[1.5pt]
\makecell[c]{\textbf{T5} Liquid Pouring} &
$66.00\pm6.00$ & $59.33\pm6.43$ & $52.67\pm9.02$ & $58.67\pm3.06$ & $57.33\pm7.57$ & $\mathbf{72.67}\pm5.03$\\
\addlinespace[1.5pt]
\makecell[c]{\textbf{T6} Block Stacking} &
$68.00\pm9.17$ & $66.67\pm7.02$ & $60.67\pm9.87$ & $62.67\pm8.33$ & $58.00\pm9.17$ & $\mathbf{71.33}\pm2.31$\\
\midrule
\textbf{Overall} &
$74.22\pm1.17$ & $69.44\pm3.10$ & $65.11\pm3.75$ & $67.78\pm2.04$ & $66.11\pm1.71$ & $\mathbf{77.67}\pm1.76$\\
\bottomrule
\end{tabular*}
\end{table*}

\subsection{Ablation Studies}\label{sec:experiments_ablation}

We next isolate the two central design decisions of the DriftingVLA framework: the geometry used to organize action chunks for drifting, and the way a pretrained VLA is adapted to the new direct generative objective.

\paragraph{Action-chunk drifting geometry}
Table~\ref{tab:ablation}, panel (a) compares PDTD with the Chunk-wise and Step-wise geometries considered in DBP under the same DriftingVLA training framework. PDTD improves over Chunk-wise drifting by 3.04 points on LIBERO and 6.92 points on RoboTwin~2.0, and exceeds Step-wise drifting by 5.22 and 10.66 points, respectively. The larger separation on RoboTwin~2.0 shows that the benefit of PDTD remains pronounced on a broader bimanual benchmark with domain randomization. These results support the structural analysis in Sec.~\ref{sec:method_geometry}: sharing one geometry across the entire heterogeneous action chunk can introduce cross-channel interference, whereas splitting by timestep fragments the temporal trajectory of each action channel. PDTD retains the full temporal trajectory while allowing channel-specific geometry. The same Step--Chunk--PDTD ordering in the real-world results of Table~\ref{tab:realworld} further supports the transfer of this geometric advantage beyond simulation.

\paragraph{Adapting a pretrained VLA to drifting}
Table~\ref{tab:ablation}, panel (b) evaluates the main choices introduced in Sec.~\ref{sec:method_formulation}. First, simply evaluating the trained $\pi_{0.5}$ flow policy with one NFE reduces LIBERO performance from $97.10\%$ to $94.02\%$, showing that solver truncation alone does not produce a native one-step generator. Second, warm-starting DriftingVLA from the flow-pretrained action expert reaches only $93.33\%$, compared with $98.32\%$ when the action interface is freshly initialized. This supports relearning the generative action interface rather than inheriting parameters specialized for local flow-velocity estimation. Third, freezing the VLM and training only the fresh action expert yields $94.57\%$. Jointly post-training the pretrained VLM and the new action expert improves success by an additional 3.75 points over the Expert-only variant, showing the importance of adapting the multimodal representation to the direct drifting objective.

\begin{table}[!t]
\centering
\caption{Controlled ablations of DriftingVLA (success rate, \%; mean \(\pm\) sample standard deviation over three seeds). (a) Action-chunk drifting geometry on LIBERO and RoboTwin~2.0. (b) Pretrained VLA adaptation on LIBERO.}\label{tab:ablation}
\scriptsize
\renewcommand{\arraystretch}{1.08}
\setlength{\tabcolsep}{3.2pt}

\textbf{(a) Action-chunk drifting geometry}\par\vspace{0.4mm}
\begin{tabular*}{\columnwidth}{@{\extracolsep{\fill}}lccc@{}}
\toprule
Geometry & NFE & LIBERO & RoboTwin~2.0\\
\midrule
Chunk-wise & 1 & $95.28\pm0.03$ & $74.17\pm0.27$\\
Step-wise & 1 & $93.10\pm0.40$ & $70.43\pm0.28$\\
\textbf{PDTD (ours)} & 1 & $\mathbf{98.32}\pm0.13$ & $\mathbf{81.09}\pm0.38$\\
\bottomrule
\end{tabular*}

\par\vspace{1.4mm}
\textbf{(b) Pretrained-VLA adaptation on LIBERO}\par\vspace{0.4mm}
\begin{tabular*}{\columnwidth}{@{\extracolsep{\fill}}lcccc@{}}
\toprule
Variant & \makecell{Expert\\init.} & \makecell{VLM\\updated} & NFE & Overall\\
\midrule
$\pi_{0.5}$ & \makecell{Pretrained\\$\pi_{0.5}$} & Yes & 10 & $97.10\pm0.13$\\
$\pi_{0.5}$-1step & \makecell{Pretrained\\$\pi_{0.5}$} & Yes & 1 & $94.02\pm0.13$\\
Warm-start & \makecell{Flow-\\pretrained} & Yes & 1 & $93.33\pm0.29$\\
Expert-only & Fresh & No & 1 & $94.57\pm0.15$\\
\textbf{DriftingVLA} & Fresh & Yes & 1 & $\mathbf{98.32}\pm0.13$\\
\bottomrule
\end{tabular*}
\end{table}

\subsection{Computational Efficiency}\label{sec:exp_efficiency}

The main motivation for native one-step generation is to remove repeated action-expert evaluations from the online control loop. We therefore examine DriftingVLA from both deployment-time efficiency and training-time overhead.

\paragraph{Inference latency}
We profile the seed-42, 50k-step checkpoints on LIBERO-Spatial task~0 with batch size 1 and $H=50$ on a single NVIDIA A800 GPU. Each method is measured over 10 episodes and three independent profiling repetitions, with three warm-up chunk generations excluded from each run. CUDA events are used for model-stage timing and synchronized wall-clock measurements for the surrounding pipeline.

\begin{figure}[!t]
    \centering
    \includegraphics[width=\columnwidth]{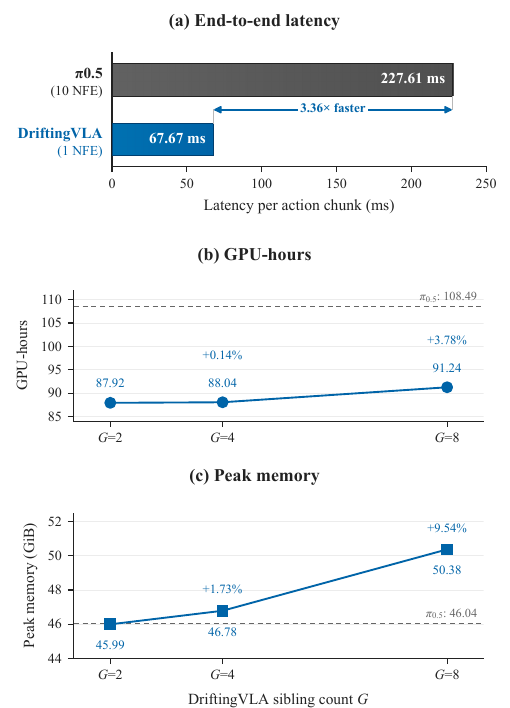}
    \caption{Computational efficiency of DriftingVLA. (a) End-to-end action-chunk inference latency for 10-NFE \(\pi_{0.5}\) and 1-NFE DriftingVLA. (b,c) Training GPU-hours and peak memory of DriftingVLA as the number of drifting siblings increases from \(G=2\) to \(G=8\) under a fixed profiling setup. Dashed lines indicate the corresponding \(\pi_{0.5}\) references, and percentage annotations report changes relative to \(G=2\).}
    \label{fig:efficiency}
\end{figure}

As shown in Fig.~\ref{fig:efficiency}(a), DriftingVLA reduces mean action-chunk latency from 227.61\,ms to 67.67\,ms, corresponding to a $3.36\times$ speedup and a 70.3\% reduction in latency. This reduction follows directly from replacing the 10-NFE iterative inference of $\pi_{0.5}$ with a single generator evaluation, removing repeated action-expert evaluations from the deployment-time control loop.

\paragraph{Training cost}
Native one-step inference shifts distribution refinement to training, where DriftingVLA uses multiple sibling samples to estimate the drifting field. We therefore examine how this additional sampling affects optimization cost. Using LIBERO with seed 43, $H=50$, 50k steps, 8 NVIDIA A800 GPUs, and a global batch size of 32, we profile DriftingVLA with $G\in\{2,4,8\}$ under an otherwise fixed training configuration; checkpoint saving and evaluation are disabled so that GPU-hours reflect the training loop itself.

Figures~\ref{fig:efficiency}(b,c) show that increasing the number of siblings introduces only moderate training overhead. Increasing $G$ from 2 to 8 raises the measured training cost from 87.92 to 91.24 GPU-hours, an increase of 3.78\%, while peak memory grows from 45.99 to 50.38\,GiB, corresponding to 9.54\%. At the default $G=8$, DriftingVLA requires 91.24 GPU-hours under the controlled profiling setup, remaining below the corresponding $\pi_{0.5}$ reference of 108.49 GPU-hours. The mild scaling with $G$ reflects the shared-context sampling design in Sec.~\ref{sec:method_formulation}: the expensive VLM context is computed once per condition, while additional siblings primarily expand action-side computation.

\paragraph{Summary}
Overall, the experiments support the central design choices of
DriftingVLA: native one-step generation preserves or improves control
performance, PDTD outperforms the DBP geometries, and the resulting policy
substantially reduces deployment-time latency with moderate training
overhead.

\section{Conclusion and Limitations}
\label{sec:conclusion}

This work introduced DriftingVLA, a native one-step VLA that replaces inference-time iterative flow transport with training-time distribution drifting. Starting from pretrained $\pi_{0.5}$ multimodal representations, DriftingVLA relearns the continuous action interface as a direct conditional generator and introduces Per-Dimension Temporal Drifting (PDTD), which defines channel-specific drifting geometry over complete temporal trajectories while preserving joint action generation. Across LIBERO, RoboTwin~2.0, and six real-world single- and dual-arm manipulation tasks, DriftingVLA outperforms the original 10-NFE $\pi_{0.5}$ policy and representative one-step baselines in overall success while reducing action-chunk latency from 227.61\,ms to 67.67\,ms, corresponding to a $3.36\times$ speedup. The ablations further show the importance of PDTD, fresh action-interface learning, and joint VLM adaptation. Together, these results show that native one-step VLA generation can remove iterative action refinement from deployment without sacrificing control performance.

DriftingVLA uses multiple sibling samples during training, introducing
additional computation and memory, although this overhead is absent at
deployment. Our evaluation also focuses on manipulation tasks and adopts a fixed per-dimension temporal grouping for the action space. Future work can explore adaptive or learned drifting geometries, extend native one-step generation to larger multi-embodiment VLAs with heterogeneous action spaces, and evaluate the approach in broader long-horizon and interactive robotic settings.

\bibliographystyle{IEEEtran}
\bibliography{ref}

@inproceedings{pi0_2024,
  author    = {Kevin Black and Noah Brown and Danny Driess and Adnan Esmail and Michael Robert Equi and Chelsea Finn and Niccolo Fusai and Lachy Groom and Karol Hausman and Brian Ichter and Szymon Jakubczak and Tim Jones and Liyiming Ke and Sergey Levine and Adrian Li-Bell and Mohith Mothukuri and Suraj Nair and Karl Pertsch and Lucy Xiaoyang Shi and Laura Smith and James Tanner and Quan Vuong and Anna Walling and Haohuan Wang and Ury Zhilinsky},
  title     = {{$\pi_0$: A Vision-Language-Action Flow Model for General Robot Control}},
  booktitle = {Proceedings of Robotics: Science and Systems},
  year      = {2025},
  doi       = {10.15607/RSS.2025.XXI.010}
}

@article{pi05_2024,
  title={${\pi}_{0.5}$: A Vision-Language-Action Model with Open-World Generalization},
  author={{Physical Intelligence} and Black, Kevin and Brown, Noah and Darpinian, James and Dhabalia, Karan and Driess, Danny and others},
  journal={arXiv preprint arXiv:2504.16054},
  year={2025}
}

@inproceedings{lipman2023flow,
  title={Flow Matching for Generative Modeling},
  author={Lipman, Yaron and Chen, Ricky T. Q. and Ben-Hamu, Heli and Nickel, Maximilian and Le, Matthew},
  booktitle={International Conference on Learning Representations},
  year={2023}
}

@article{deng2026generative,
  title={Generative Modeling via Drifting},
  author={Deng, Mingyang and Li, He and Li, Tianhong and Du, Yilun and He, Kaiming},
  journal={arXiv preprint arXiv:2602.04770},
  year={2026}
}

@inproceedings{gao2026drift,
  author    = {Yuxuan Gao and Yedong Shen and Shiqi Zhang and
               Wenhao Yu and Yifan Duan and Jia Pan and Jiajia Wu and
               Jiajun Deng and Yanyong Zhang},
  title     = {Drift-Based Policy Optimization: Native One-Step Policy
               Learning for Online Robot Control},
  booktitle = {Proceedings of the 34th ACM International Conference
               on Multimedia},
  year      = {2026},
  doi       = {10.1145/3767308.3836199},
  note      = {To appear; preprint available at arXiv:2604.03540}
}

@inproceedings{meanflow2024,
  title={Mean Flows for One-step Generative Modeling},
  author={Geng, Zhengyang and Deng, Mingyang and Bai, Xingjian and Kolter, J. Zico and He, Kaiming},
  booktitle={Advances in Neural Information Processing Systems},
  volume={38},
  year={2025}
}

@inproceedings{yan2025maniflow,
  title={ManiFlow: A General Robot Manipulation Policy via Consistency Flow Training},
  author={Yan, Ge and Zhu, Jiyue and Deng, Yuquan and Yang, Shiqi and Qiu, Ri-Zhao and Cheng, Xuxin and Memmel, Marius and Krishna, Ranjay and Goyal, Ankit and Wang, Xiaolong and Fox, Dieter},
  booktitle={Proceedings of The 9th Conference on Robot Learning},
  series={Proceedings of Machine Learning Research},
  volume={305},
  pages={2268--2293},
  year={2025}
}

@inproceedings{prasad2024consistency,
  title={Consistency Policy: Accelerated Visuomotor Policies via Consistency Distillation},
  author={Prasad, Aaditya and Lin, Kevin and Wu, Jimmy and Zhou, Linqi and Bohg, Jeannette},
  booktitle={Robotics: Science and Systems},
  year={2024}
}

@inproceedings{wang2024onestepdiffusionpolicyfast,
  title={One-Step Diffusion Policy: Fast Visuomotor Policies via Diffusion Distillation},
  author={Wang, Zhendong and Li, Max and Mandlekar, Ajay and Xu, Zhenjia and Fan, Jiaojiao and Narang, Yashraj and Fan, Linxi and Zhu, Yuke and Balaji, Yogesh and Zhou, Mingyuan and Liu, Ming-Yu and Zeng, Yu},
  booktitle={International Conference on Machine Learning},
  year={2025}
}

@article{frans2024one,
  title={One Step Diffusion via Shortcut Models},
  author={Frans, Kevin and Hafner, Danijar and Levine, Sergey and Abbeel, Pieter},
  journal={arXiv preprint arXiv:2410.12557},
  year={2024}
}

@article{sheng2025mp1,
  title={MP1: MeanFlow Tames Policy Learning in 1-step for Robotic Manipulation},
  author={Sheng, Juyi and Wang, Ziyi and Li, Peiming and Liu, Mengyuan},
  journal={arXiv preprint arXiv:2507.10543},
  year={2025}
}

@article{Fang2025OMPOM,
  title={OMP: One-step Meanflow Policy with Directional Alignment},
  author={Fang, Han and Huang, Yize and Zhao, Yuheng and Weng, Paul and Li, Xiao and Ban, Yutong},
  journal={arXiv preprint arXiv:2512.19347},
  year={2025}
}

@article{snapflow2026,
  title={SnapFlow: One-Step Action Generation for Flow-Matching VLAs via Progressive Self-Distillation},
  author={Luan, Wuyang and Li, Junhui and Zhao, Weiguang and Zhang, Wenjian and Wu, Tieru and Ma, Rui},
  journal={arXiv preprint arXiv:2604.05656},
  year={2026}
}

@article{meanflowvla_2026,
  title={Mean-Flow based One-Step Vision-Language-Action},
  author={Chen, Yang and Ma, Xiaoguang and Zhao, Bin},
  journal={arXiv preprint arXiv:2603.01469},
  year={2026}
}

@inproceedings{libero2023,
  title={LIBERO: Benchmarking Knowledge Transfer for Lifelong Robot Learning},
  author={Liu, Bo and Zhu, Yifeng and Gao, Chongkai and Feng, Yihao and Liu, Qiang and Zhu, Yuke and Stone, Peter},
  booktitle={Advances in Neural Information Processing Systems},
  year={2023}
}

@article{robotwin2025,
  title={RoboTwin 2.0: A Scalable Data Generator and Benchmark with Strong Domain Randomization for Robust Bimanual Robotic Manipulation},
  author={Chen, Tianxing and Chen, Zanxin and Chen, Baijun and Cai, Zijian and Liu, Yibin and Li, Zixuan and Liang, Qiwei and Lin, Xianliang and Ge, Yiheng and Gu, Zhenyu and others},
  journal={arXiv preprint arXiv:2506.18088},
  year={2025}
}

@inproceedings{rt2_2023,
  title={RT-2: Vision-Language-Action Models Transfer Web Knowledge to Robotic Control},
  author={Zitkovich, Brianna and Yu, Tianhe and Xu, Sichun and Xu, Peng and Xiao, Ted and Xia, Fei and others},
  booktitle={Proceedings of The 7th Conference on Robot Learning},
  series={Proceedings of Machine Learning Research},
  volume={229},
  pages={2165--2183},
  year={2023}
}

@inproceedings{octo2024,
  title={Octo: An Open-Source Generalist Robot Policy},
  author={Ghosh, Dibya and Walke, Homer Rich and Pertsch, Karl and Black, Kevin and Mees, Oier and Dasari, Sudeep and others},
  booktitle={Proceedings of Robotics: Science and Systems},
  year={2024}
}

@inproceedings{openvla2025,
  title={OpenVLA: An Open-Source Vision-Language-Action Model},
  author={Kim, Moo Jin and Pertsch, Karl and Karamcheti, Siddharth and Xiao, Ted and Balakrishna, Ashwin and Nair, Suraj and others},
  booktitle={Proceedings of The 8th Conference on Robot Learning},
  series={Proceedings of Machine Learning Research},
  volume={270},
  pages={2679--2713},
  year={2025}
}

@inproceedings{fast_2025,
  author    = {Karl Pertsch and Kyle Stachowicz and Brian Ichter and Danny Driess and Suraj Nair and Quan Vuong and Oier Mees and Chelsea Finn and Sergey Levine},
  title     = {{FAST: Efficient Action Tokenization for Vision-Language-Action Models}},
  booktitle = {Proceedings of Robotics: Science and Systems},
  year      = {2025},
  doi       = {10.15607/RSS.2025.XXI.012}
}

@inproceedings{rtc_2025,
  title={Real-Time Execution of Action Chunking Flow Policies},
  author={Black, Kevin and Galliker, Manuel Y. and Levine, Sergey},
  booktitle={Advances in Neural Information Processing Systems},
  year={2025}
}

@article{faster_2026,
  title={FASTER: Rethinking Real-Time Flow VLAs},
  author={Lu, Yuxiang and Liu, Zhe and Fan, Xianzhe and Yang, Zhenya and Hou, Jinghua and Li, Junyi and Ding, Kaixin and Zhao, Hengshuang},
  journal={arXiv preprint arXiv:2603.19199},
  year={2026}
}

@article{pi0eqm_2026,
  title={${\pi}_0$-EqM: Equilibrium Matching for Closed-Loop Vision-Language-Action Control},
  author={Liu, Huanming and Xu, Congsheng and Ji, Jianmin and Mu, Yao},
  journal={arXiv preprint arXiv:2605.23128},
  year={2026}
}

@inproceedings{diffusionpolicy2023,
  title={Diffusion Policy: Visuomotor Policy Learning via Action Diffusion},
  author={Chi, Cheng and Feng, Siyuan and Du, Yilun and Xu, Zhenjia and Cousineau, Eric and Burchfiel, Benjamin and Song, Shuran},
  booktitle={Proceedings of Robotics: Science and Systems},
  year={2023}
}

@inproceedings{rt1_2023,
  title={RT-1: Robotics Transformer for Real-World Control at Scale},
  author={Brohan, Anthony and Brown, Noah and Carbajal, Justice and
          Chebotar, Yevgen and Dabis, Joseph and Finn, Chelsea and others},
  booktitle={Robotics: Science and Systems},
  year={2023}
}

@inproceedings{openx2024,
  title={Open X-Embodiment: Robotic Learning Datasets and RT-X Models},
  author={{Open X-Embodiment Collaboration} and O'Neill, Abby and
          Rehman, Abdul and Gupta, Abhinav and Maddukuri, Abhiram and others},
  booktitle={IEEE International Conference on Robotics and Automation},
  year={2024},
  doi={10.1109/ICRA57147.2024.10611477}
}

@inproceedings{act2023,
  title={Learning Fine-Grained Bimanual Manipulation with Low-Cost Hardware},
  author={Zhao, Tony Z. and Kumar, Vikash and Levine, Sergey and Finn, Chelsea},
  booktitle={Robotics: Science and Systems},
  year={2023},
  doi={10.15607/RSS.2023.XIX.016}
}

@inproceedings{palme2023,
  title={PaLM-E: An Embodied Multimodal Language Model},
  author={Driess, Danny and Xia, Fei and Sajjadi, Mehdi S. M. and Lynch, Corey and Chowdhery, Aakanksha and Ichter, Brian and Wahid, Ayzaan and Tompson, Jonathan and Vuong, Quan and Yu, Tianhe and Huang, Wenlong and Chebotar, Yevgen and Sermanet, Pierre and Duckworth, Daniel and Levine, Sergey and Vanhoucke, Vincent and Hausman, Karol and Toussaint, Marc and Greff, Klaus and Zeng, Andy and Mordatch, Igor and Florence, Pete},
  booktitle={Proceedings of the 40th International Conference on Machine Learning},
  series={Proceedings of Machine Learning Research},
  volume={202},
  pages={8469--8488},
  year={2023},
}

@article{openvlaoft2025,
  title={Fine-Tuning Vision-Language-Action Models: Optimizing Speed and Success},
  author={Kim, Moo Jin and Finn, Chelsea and Liang, Percy},
  journal={arXiv preprint arXiv:2502.19645},
  year={2025},
}

@article{smolvla2025,
  title={SmolVLA: A Vision-Language-Action Model for Affordable and Efficient Robotics},
  author={Shukor, Mustafa and Aubakirova, Dana and Capuano, Francesco and Kooijmans, Pepijn and Palma, Steven and Zouitine, Adil and Aractingi, Michel and Pascal, Caroline and Russi, Martino and Marafioti, Andres and Alibert, Simon and Cord, Matthieu and Wolf, Thomas and Cadene, Remi},
  journal={arXiv preprint arXiv:2506.01844},
  year={2025},
}

@article{rdt1b2024,
  title={RDT-1B: A Diffusion Foundation Model for Bimanual Manipulation},
  author={Liu, Songming and Wu, Lingxuan and Li, Bangguo and Tan, Hengkai and Chen, Huayu and Wang, Zhengyi and Xu, Ke and Su, Hang and Zhu, Jun},
  journal={arXiv preprint arXiv:2410.07864},
  year={2024},
}

@article{cogact2024,
  title={CogACT: A Foundational Vision-Language-Action Model for Synergizing Cognition and Action in Robotic Manipulation},
  author={Li, Qixiu and Liang, Yaobo and Wang, Zeyu and Luo, Lin and Chen, Xi and Liao, Mozheng and Wei, Fangyun and Deng, Yu and Xu, Sicheng and Zhang, Yizhong and Wang, Xiaofan and Liu, Bei and Fu, Jianlong and Bao, Jianmin and Chen, Dong and Shi, Yuanchun and Yang, Jiaolong and Guo, Baining},
  journal={arXiv preprint arXiv:2411.19650},
  year={2024},
}

@inproceedings{baku2024,
  title={BAKU: An Efficient Transformer for Multi-Task Policy Learning},
  author={Haldar, Siddhant and Peng, Zhuoran and Pinto, Lerrel},
  booktitle={Advances in Neural Information Processing Systems},
  volume={37},
  year={2024},
  doi={10.52202/079017-4484}
}

@article{rectifiedflow2022,
  title={Flow Straight and Fast: Learning to Generate and Transfer Data with Rectified Flow},
  author={Liu, Xingchao and Gong, Chengyue and Liu, Qiang},
  journal={arXiv preprint arXiv:2209.03003},
  year={2022},
}

@article{asyncvla2025,
  title={AsyncVLA: Asynchronous Flow Matching for Vision-Language-Action Models},
  author={Jiang, Yuhua and Cheng, Shuang and Ding, Yan and Gao, Feifei and Qi, Biqing},
  journal={arXiv preprint arXiv:2511.14148},
  year={2025},
}

@inproceedings{falcon2025,
  title={Falcon: Fast Visuomotor Policies via Partial Denoising},
  author={Chen, Haojun and Liu, Minghao and Ma, Chengdong and Ma, Xiaojian and Ma, Zailin and Wu, Huimin and Chen, Yuanpei and Zhong, Yifan and Wang, Mingzhi and Li, Qing and Yang, Yaodong},
  booktitle={Proceedings of the 42nd International Conference on Machine Learning},
  series={Proceedings of Machine Learning Research},
  volume={267},
  pages={8552--8573},
  year={2025},
}

@article{letitbesimple2026,
  title={Let It Be Simple: One-Step Action Generation for Vision-Language-Action Models},
  author={Chen, Yitong and Zhang, Shiduo and Gong, Jingjing and Qiu, Xipeng},
  journal={arXiv preprint arXiv:2606.05737},
  year={2026},
}

@article{invertibleadapter2026,
  title={Invertible Neural Network Adapter for One-Step Flow Matching in Robot Manipulation},
  author={Zhang, Yu and Ji, Kangyi and Zou, Yongxiang and Xu, Rongtao and Zheng, Feng and Cheng, Long},
  journal={arXiv preprint arXiv:2606.19194},
  year={2026},
}

@article{mmact2025,
  title={MM-ACT: Learn from Multimodal Parallel Generation to Act},
  author={Liang, Haotian and Chen, Xinyi and Wang, Bin and Chen, Mingkang and Liu, Yitian and Zhang, Yuhao and Chen, Zanxin and Yang, Tianshuo and Chen, Yilun and Pang, Jiangmiao and Liu, Dong and Yang, Xiaokang and Mu, Yao and Shao, Wenqi and Luo, Ping},
  journal={arXiv preprint arXiv:2512.00975},
  year={2025},
}

@article{reactvla2026,
  title={ReactVLA: Fast and Lightweight Reactive Robot Manipulation via Improved Mean Flow Action Generation},
  author={Guo, Yanzhao and Chen, Wenkai and Zhang, Jianwei},
  journal={arXiv preprint arXiv:2606.14255},
  year={2026},
}

@article{grootn1_2025,
  title={GR00T N1: An Open Foundation Model for Generalist Humanoid Robots},
  author={{NVIDIA} and Bjorck, Johan and Castañeda, Fernando and Cherniadev, Nikita and Da, Xingye and Ding, Runyu and Fan, Linxi and Fang, Yu and Fox, Dieter and Hu, Fengyuan and Huang, Spencer and Jang, Joel and Jiang, Zhenyu and Kautz, Jan and Kundalia, Kaushil and Lao, Lawrence and Li, Zhiqi and Lin, Zongyu and Lin, Kevin and Liu, Guilin and Llontop, Edith and Magne, Loic and Mandlekar, Ajay and Narayan, Avnish and Nasiriany, Soroush and Reed, Scott and Tan, You Liang and Wang, Guanzhi and Wang, Zu and Wang, Jing and Wang, Qi and Xiang, Jiannan and Xie, Yuqi and Xu, Yinzhen and Xu, Zhenjia and Ye, Seonghyeon and Yu, Zhiding and Zhang, Ao and Zhang, Hao and Zhao, Yizhou and Zheng, Ruijie and Zhu, Yuke},
  journal={arXiv preprint arXiv:2503.14734},
  year={2025},
}

@article{arvla2026,
  title={AR-VLA: True Autoregressive Action Expert for Vision-Language-Action Models},
  author={Hu, Yutong and Zaech, Jan-Nico and Nikolov, Nikolay and Yao, Yuanqi and Dey, Sombit and Albanese, Giuliano and Detry, Renaud and Van Gool, Luc and Paudel, Danda},
  journal={arXiv preprint arXiv:2603.10126},
  year={2026},
  note={Accepted to Robotics: Science and Systems 2026},
}

@article{onestepflowpolicy2026,
  title={One-Step Flow Policy: Self-Distillation for Fast Visuomotor Policies},
  author={Li, Shaolong and Sun, Lichao and Chen, Yongchao},
  journal={arXiv preprint arXiv:2603.12480},
  year={2026},
}

@inproceedings{dp3_2024,
  title={3D Diffusion Policy: Generalizable Visuomotor Policy Learning via Simple 3D Representations},
  author={Ze, Yanjie and Zhang, Gu and Zhang, Kangning and Hu, Chenyuan and Wang, Muhan and Xu, Huazhe},
  booktitle={Proceedings of Robotics: Science and Systems},
  year={2024},
  doi={10.15607/RSS.2024.XX.067},
}

\end{document}